\documentclass[letterpaper, 10 pt, conference]{ieeeconf}

\IEEEoverridecommandlockouts

\usepackage{etextools}
\usepackage{amssymb}
\usepackage{float}
\usepackage{makeidx}
\usepackage{amsmath}
\usepackage{bbm}
\usepackage{epsfig}
\usepackage{epsf}
\usepackage{psfrag}
\usepackage{verbatim}
\usepackage{color}
\usepackage{multirow}
\usepackage{tabularx}
\usepackage{booktabs}
\usepackage[tight,footnotesize]{subfigure}
\usepackage{array}
\usepackage{soul}
\usepackage{footnote}
\usepackage{cite}
\usepackage{dblfloatfix}

\usepackage{xcolor}
\usepackage{color, colortbl}
\usepackage[colorlinks,bookmarksnumbered,citecolor=orange,urlcolor=orange]{hyperref}
\usepackage{graphicx}
\graphicspath{{Figures}}
\DeclareGraphicsExtensions{.pdf,.png}
\usepackage{bigstrut}
\usepackage[english]{babel}

\usepackage{csquotes}

\usepackage{algorithm, setspace}
\usepackage{algpseudocode}
\usepackage{url}
\usepackage{multirow}
\usepackage{float}
 \hypersetup{
     colorlinks=true,
     linkcolor=orange,
     filecolor=orange,
     citecolor=orange,      
     urlcolor=orange,
     }
\usepackage[T1]{fontenc}
\usepackage{balance}

\newcommand{\p}[1]{\smallskip \noindent \textbf{{#1}.}}
\newcommand{\eq}[1]{Equation~(\ref{eq:#1})}
\newcommand{\fig}[1]{Figure~\ref{fig:#1}}

\title{\LARGE

Learning to Share Autonomy Across Repeated Interaction

}

\author{Ananth Jonnavittula and Dylan P. Losey
\thanks{The authors are members of the Collaborative Robotics Lab (\href{https://collab.me.vt.edu/}{Collab}), Dept. of Mechanical Engineering, Virginia Tech, Blacksburg, VA 24061.
\newline
{e-mail: \texttt{\{ananth, losey\}@vt.edu}}}
}

\begin{document}
\maketitle

\begin{abstract}

Wheelchair-mounted robotic arms (and other assistive robots) should help their users perform everyday tasks. One way robots can provide this assistance is \textit{shared autonomy}. Within shared autonomy, both the human and robot maintain control over the robot’s motion: as the robot becomes confident it understands what the human wants, it increasingly intervenes to automate the task. But how does the robot know what tasks the human may want to perform in the first place? Today’s shared autonomy approaches often rely on \textit{prior knowledge}: for example, the robot must know the set of possible human goals \textit{a priori}. In the long-term, however, this prior knowledge will inevitably break down --- sooner or later the human will reach for a goal that the robot did not expect. In this paper we propose a learning approach to shared autonomy that takes advantage of repeated interactions. Learning to assist humans would be impossible if they performed completely different tasks at every interaction: but our insight is that users living with physical disabilities \textit{repeat} important tasks on a daily basis (e.g., opening the fridge, making coffee, and having dinner). We introduce an algorithm that exploits these repeated interactions to \textit{recognize} the human’s task, \textit{replicate} similar demonstrations, and \textit{return} control when unsure. As the human repeatedly works with this robot, our approach continually learns to assist tasks that were never specified beforehand: these tasks include both discrete goals (e.g., reaching a cup) and continuous skills (e.g., opening a drawer). Across simulations and an in-person user study, we demonstrate that robots leveraging our approach match existing shared autonomy methods for known goals, and outperform imitation learning baselines on new tasks. See videos here: \url{https://youtu.be/Plh4t3wQeIA}

\end{abstract}


\section{Introduction}

\begin{figure}[t!]
	\begin{center}
		\includegraphics[width=0.8\columnwidth]{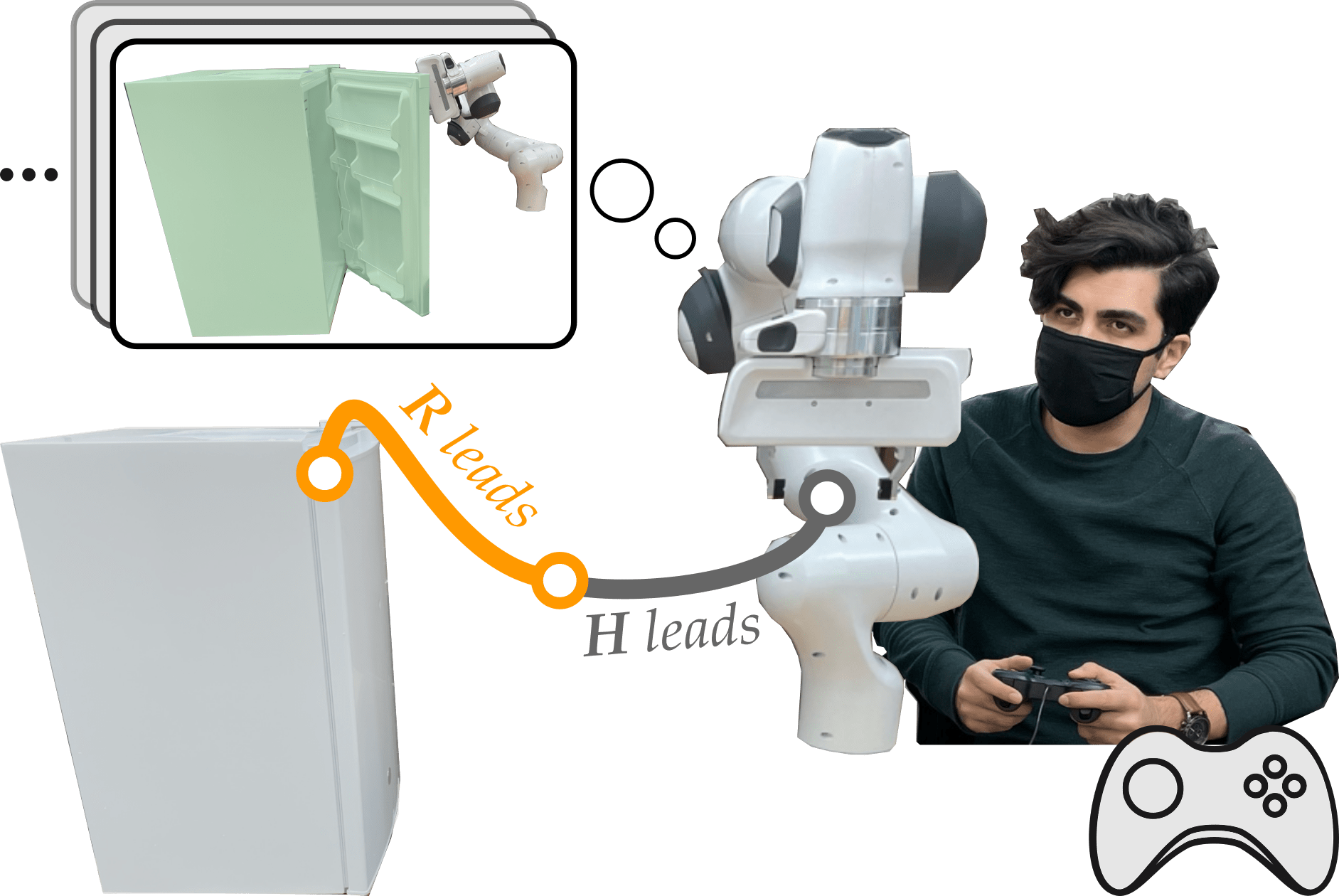}

		\caption{User teleoperating an assistive robot arm to open their fridge door. The robot does not have any pre-programmed knowledge about this task; however, the human and robot have completed similar tasks many times before. We enable robots to learn to assist humans from scratch by exploiting the \textit{repeated} nature of everyday tasks.}
		\label{fig:front}
	\end{center}
	\vspace{-2em}

\end{figure}

Imagine teleoperating a wheelchair-mounted robot arm to open your refrigerator door (see \fig{front}). The first few times you need to open the fridge, you must carefully guide the robot throughout the entire process of reaching, pulling, and opening the door. But after you've interacted with this robot for several weeks --- and opened your fridge \textit{many} times --- the intelligent robot should learn to \textit{assist} you. The next time you start teleoperating the arm towards your fridge, this robot should recognize what you want, and autonomously take over to help pull open the door.

Today's assistive robot arms often leverage \textit{shared autonomy} to help their users perform complex tasks \cite{dragan2013policy, jain2019probabilistic, javdani2018shared, brooks2019balanced, gopinath2016human, newman2018harmonic, nikolaidis2017human}. Here the robot is given a discrete set of tasks the human may want to complete \textit{a priori}. For instance, the human may want to reach for their cup or move their plate. Based on the human's inputs so far, the robot infers which task the human is trying to perform, and arbitrates control to automate that task.

This approach to shared autonomy makes sense when the robot knows \textit{all} your potential tasks --- but what happens when some tasks are inevitably \textit{left out}? Going back to our example, let's say the robot does not know you might want to open the fridge. As you guide the arm towards the door, the robot gets confused: this robot is unsure about what task you are trying to do, and cannot assist you in this unexpected task. Even worse, the robot remains confused no matter how many times you repeat the process of opening the fridge.

For assistive robot arms to be helpful in practice, these robots must be capable of \textit{learning} a spectrum of new tasks over time. This would be extremely challenging if every task was a unique one-off the robot had never seen before. But our insight is that, over the many weeks, months, and years a user living with disabilities works with their assistive robot:
\begin{center}\vspace{-0.3em}
\textit{Humans constantly} repeat \textit{tasks\\ that are important in their everyday life.}\vspace{-0.3em}
\end{center}
For example, you open your refrigerator door on a daily basis. Applying our insight enables assistive robot arms to learn to share autonomy by exploiting the \textit{repeated interaction} inherent in assistive applications. Here the robot remembers how you controlled the arm to open the fridge in the past, recognizes that you are providing similar inputs during the current interaction, and assists by autonomously mimicking the behavior you previously demonstrated. Across repeated interactions, these robot arms should learn to assist not only \textit{discrete goals} (e.g., reaching a cup) but also \textit{continuous skills} (e.g., opening a door).

Overall, we make the following contributions:

\p{Capturing Latent Intent} We formalize shared autonomy with repeated interaction. During each interaction the human has in mind some desired task: we introduce an autoencoder approach that learns to recognize the human's current task and replicate similar past interactions.

\p{Returning Control when Uncertain} Our approach should assist during previously seen tasks without interfering whenever the human tries to do a new task. We introduce a discriminator to measure the confidence of our learned shared autonomy, so that the robot returns control to the human whenever it is unsure about what the human really wants.

\p{Conducting a User Study} We assess our resulting algorithm in scenarios where the human is trying to reach discrete goals and perform continuous skills. We baseline against safe imitation learning and shared autonomy approaches, and show that our algorithm enables the robot to learn to assist for initially unknown tasks across repeated interactions.

\section{Related Work}

\p{Application -- Assistive Robot Arms} Over $13\%$ of American adults living with physical disabilities have difficulty with at least one activity of daily living (ADL) \cite{taylor2018americans}. Assistive robots --- such as wheelchair-mounted robot arms \cite{argall2018autonomy} --- have the potential to help users perform these everyday tasks without relying on caregivers. Recent work on assistive robot arms has focused on automating ADLs such as eating dinner \cite{feng2019robot, jeon2020shared, park2020active}, getting dressed \cite{erickson2020assistive}, and reaching common objects \cite{choi2009list}. Our research takes inspiration from the fact that people living with disabilities need \textit{assistance} performing tasks that they \textit{repeat} on a daily basis.

\p{Shared Autonomy} How should we provide this assistance? Rather than relying on the human to constantly teleoperate the robot arm, often it makes sense for the robot to automate parts of the task it understands. In shared autonomy both the human and robot arbitrate control over the robot's motion. We divide related work on shared autonomy into two groups: \textit{inference} and \textit{optimization}.

Within inference works such as \cite{dragan2013policy, jain2019probabilistic, javdani2018shared, brooks2019balanced, gopinath2016human, newman2018harmonic, nikolaidis2017human}, the robot is given a discrete set of possible goals the human may want to reach \textit{a priori}. Based on the human's inputs so far, the robot infers which goal is most likely, and autonomously moves towards that target. The level of robotic assistance is proportional to the robot's confidence in the human's goal.

By contrast, within optimization works like \cite{bragg2020fake, broad2020data, reddy2018shared, schaff2020residual} the robot is given an underlying reward function that the human wants to optimize (e.g., maximize distance from obstacles). At timesteps when the user's input would lead to low reward, the robot intervenes, and corrects the human's action.

Both approaches to shared autonomy require \textit{prior knowledge} about what the human wants: the robot must observe the human's potential goals or underlying reward function. But in the long-term these priors will inevitably fail --- sooner or later the human will reach for a goal that the robot did not expect. Accordingly, here we remove this key assumption by exploiting \textit{repetition} in assistive human-robot interaction.

\p{Repeated Interaction} Prior work enables robots to adapt to humans over the course of repeated interaction. For example, robots can recognize emerging conventions in human inputs \cite{shih2020critical}, or imitate the human's behavior \cite{ross2011reduction}. Most relevant to our approach is existing research on \textit{latent representations} \cite{xie2020learning, lynch2020learning, jonschkowski2015learning,  pertsch2020accelerating}. Here the robot learns to embed complex behavior into a compact, high-level representation (e.g., skills, strategies, or plans). We similarly apply latent representations to learn the human's desired \textit{tasks} across repeated interaction.
\section{Formalizing Shared Autonomy with Repeated Interaction}

\begin{figure*}[t!]
	\begin{center}
		\includegraphics[width=2\columnwidth]{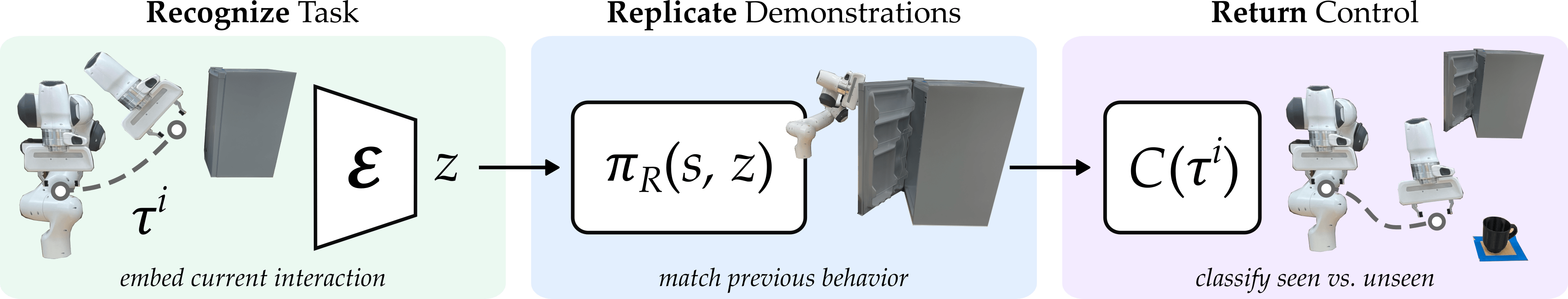}

		\caption{Our proposed approach for learning to share autonomy across repeated interaction. (Left) The robot embeds the human's behavior $\tau^i$ during the current interaction to a distribution over latent tasks $z$. (Middle) The robot then chooses assistive actions $a_{\mathcal{R}}$ conditioned on its state $s$ and latent task $z$. Policy $\pi_{\mathcal{R}}$ is trained to match the user's behavior from previous interactions. (Right) To decide whether or not to trust this assistive action, the robot turns to a discriminator $\mathcal{C}$. The discriminator assesses whether the current interaction $\tau^i$ is similar to any previously seen interaction: if so, the robot increases autonomy. In this example the robot remembers how the human has opened the fridge in the past, and assists for that task. But when the human does something new (reaching for the cup) the robot realizes that it does not know how to help, and arbitrates control back to the human.}
		\label{fig:methods}
	\end{center}
    \vspace{-1em}

\end{figure*}

Let us return to our motivating example where the user is teleoperating their assistive robot arm. Each time the human interacts with the robot, they have in mind a \textit{task} they want the robot to perform: some of these tasks are new (e.g., moving a coffee cup), while others the robot may have seen before (e.g., opening the fridge). We represent the human's current task as $z \in \mathcal{Z}$, so that during interaction $i$, the human wants to complete task $z^i$. Within this paper tasks include both discrete goals and continuous skills: i.e., a task $z$ could be reaching the cup or opening a drawer. We test both types of tasks $z$ in our experiments. The assistive robot's goal is to help the human complete their current task. However, the robot does not know (a) \textit{which} task the human currently has in mind or (b) \textit{how} to correctly perform that task.

\p{Dynamics} The robot arm is in state $s$ and takes action $a$. Within our experiments, $s$ is the robot's joint position, $a$ is the robot's joint velocity, and the robot has dynamics:
\begin{equation} \label{eq:P1}
    s^{t+1} = s^t + \Delta t \cdot a^t
\end{equation}
The human uses a joystick to tell the robot what action to take. Let $a_{\mathcal{H}}$ be the human's \textit{commanded action} --- i.e., the joint velocity corresponding to the human's joystick input. The robot assists the human with an \textit{autonomous action} $a_{\mathcal{R}}$, so that the overall action $a$ is a linear blend of the human's joystick input and the robot's assistive guidance \cite{jain2019probabilistic,dragan2013policy}:
\begin{equation} \label{eq:P2}
    a = \beta \cdot a_{\mathcal{R}} + (1 - \beta) \cdot a_{\mathcal{H}}
\end{equation}
Here $\beta \in [0, 1]$ \textit{arbitrates} control between human and robot. When $\beta \rightarrow0$, the human always controls the robot, and when $\beta \rightarrow1$, the robot acts autonomously.

\p{Human} So how does the human choose inputs $a_{\mathcal{H}}$? During interaction $i$ we assume the human has in mind a desired task $z^i$. We know that this task guides the human's commanded actions; similar to prior work \cite{javdani2018shared}, we accordingly write the \textit{human's policy} as ${\pi_{\mathcal{H}}(a_{\mathcal{H}} \mid s, z^i)}$. This policy is the gold standard, because if we knew $\pi_h$ we would know exactly how the human likes to perform each task $z \in \mathcal{Z}$. It's important to recognize that this policy is highly \textit{personalized}. Imagine that the current task is to reach a coffee cup at state $s^*$: one human might prefer to move directly towards the cup with actions $a_{\mathcal{H}} \propto (s^* - s)$, while another user takes a circuitous route to stay farther away from obstacles.

\p{Repeated Interaction} In practice the assistive robot cannot directly observe either $z^i$ or $\pi_h$. Instead, the robot observes the states that it visits and the commands that the human provides. Let $\tau = \{(s^1, a_{\mathcal{H}}^1), \ldots, (s^T, a_{\mathcal{H}}^T)\}$ be the entire \textit{sequence} of robot states and human commands that the robot observed over the course of an interaction. As the human and robot repeatedly collaborate and interact, the robot collects a \textit{dataset} of these sequences: $\mathcal{D} = \{\tau^1, \tau^2, \ldots, \tau^{i-1} \}$. Notice that here we distinguish the current interaction $\tau^i$. Because the robot only knows the states and human inputs up to the present time, $\tau^i = \{(s^1, a_{\mathcal{H}}^1), \ldots, (s^{t-1}, a_{\mathcal{H}}^{t-1})\}$.

\p{Robot} In settings where an assistive robot arm repeatedly interacts with a human, the robot has access to three pieces of information. The robot knows its state $s$, the human's current behavior $\tau^i$, and the events of previous interactions $\mathcal{D}$. Given $(s, \tau^i, \mathcal{D})$, the robot needs to decide: (a) what assistance $a_{\mathcal{R}}$ to provide and (b) how to arbitrate control with $\beta$. We emphasize that here the robot makes no assumptions about either the human's underlying tasks or how to complete them --- instead, the robot must extract this information from previous interactions.

\section{Learning to Recognize Tasks, Replicate Interactions, and Return Control}

Our proposed approach is guided by the intuition that --- if the robot recognizes that the human's behavior is similar to a previous interaction --- the robot can assist by mimicking that previous interaction. Take our motivating example of opening the fridge door: the next time the human starts guiding the robot towards this door, the robot should infer which task the human is trying to perform, and then autonomously open the door just like the human did. There are three key challenges to this problem. First, the robot must \textbf{\textit{recognize}} the human's task $z^i$ during the current interaction. Next, the robot should \textbf{\textit{replicate}} any previous interactions that are similar to this task. Finally, the robot must know when it is unsure, and \textbf{\textit{return}} control to the human if the task is new or unexpected. In this section we outline how our approach tackles these three main challenges (see \fig{methods}).

\subsection{\emph{Recognize}: Embedding Interactions to a Latent Space} \label{sec:encoder}

Our first step is to extract the user's high-level task $z^i$ from the robot's low-level observations. Recall that the human's low-level behavior during the current interaction (i.e., their commanded actions at each robot state) is captured by $\tau^i$. This behavior is guided by the human's desired task: when the human wants to open the fridge, they provide commands $a_{\mathcal{H}}$ that move the robot towards that door, and when the human wants to pick up a coffee cup, they provide a different set of commands to reach that cup. Accordingly, we leverage $\tau^i$ to recognize the underlying task $z^i$. More formally, we introduce an encoder:
\begin{equation} \label{eq:M1}
    z \sim \mathcal{E}(~\cdot \mid \tau^i)
\end{equation}
This encoder \textit{embeds} the human's behavior to a probability distribution over the latent space $\mathcal{Z} \subseteq \mathbb{R}^d$. We learn the encoder network from previous human interactions as described in the following subsection.

Our encoder $\mathcal{E}$ is analogous to \textit{goal prediction} from prior work on shared autonomy \cite{dragan2013policy,javdani2018shared,jain2019probabilistic}. In these prior works, the robot records the human's current behavior $\tau^i$, and then applies Bayesian inference to predict the human's goal $z^i$. Our encoder ${\mathcal{E}(z \mid \tau^i)}$ practically accomplishes the same thing: it provides us with a distribution over tasks the human may want to complete. The difference is that --- when using Bayesian inference --- the robot needs to know the human's possible tasks \textit{a priori}. When training the encoder we make no such assumption. Moreover, because the sequence $\tau^i$ only considers the human's actions $a_{\mathcal{H}}$, we avoid feedback loops where the robot unintentionally uses its own actions $a_{\mathcal{R}}$ to convince itself of the human's goal \cite{javdani2018shared}.

\subsection{\emph{Replicate}: Matching the Demonstrated Behavior} \label{sec:decoder}

As the human uses their joystick to teleoperate the robot towards the fridge door, we leverage our encoder to recognize the human's task. But what does the robot do once it knows that task? And how do we train the encoder in the first place? We address both issues by introducing a decoder that maps our task predictions into assistive robot actions:
\begin{equation} \label{eq:M2}
    a_{\mathcal{R}} = \pi_{\mathcal{R}}(s, z)
\end{equation}
The decoder $\pi_{\mathcal{R}}$ determines how the robot assists the human. We want this decoder to replicate previous demonstrations, so that if the human's current behavior is similar to another interaction $\tau \in \mathcal{D}$, the robot will mimic the human's actions from that previous interaction.

We accomplish this by training the encoder and decoder models using the dataset of interactions $\mathcal{D}$. More specifically, we take snippets of the human's behavior during previous interactions, embed those snippets to a task prediction, and then reconstruct the human's demonstrated behavior. For some past interaction $\tau \in \mathcal{D}$, let $\xi = \{(s^1, a_{\mathcal{H}}^1), \ldots (s^{k-1}, a_{\mathcal{H}}^{k-1})\}$ be the human's behavior up to timestep $k$, and let $(s^k, a_{\mathcal{H}}^k)$ be the human's behavior at timestep $k$. We train the encoder and decoder to minimize the loss function:
\begin{equation} \label{eq:M3}
    \mathcal{L} = \mathbb{E}_{z \sim \mathcal{E}(\cdot \mid \xi)} ~ \|a_{\mathcal{H}}^k - \pi_{\mathcal{R}}(s^k, z) \|^2
\end{equation}
across the dataset $\mathcal{D}$. In other words, once we train $\mathcal{E}$ and $\pi_{\mathcal{R}}$, we should be able to take an a snippet of the human's past behavior and correctly predict the next action the human took. \eq{M3} encourages the robot to mimic the human, so that when we encounter a familiar task, the robot will behave like the human did.

We contrast our decoder to \textit{trajectory prediction} from prior works on shared autonomy \cite{dragan2013policy, jain2019probabilistic}. Within today's shared autonomy approaches, often the robot assumes it knows the right way to perform each task \textit{a priori}. For instance, if the human wants to reach a cup, the robot assists the human move in a straight line towards that goal. But not all humans complete tasks in the same way. Accordingly, here we \textit{learn} how the user likes to perform their tasks by replicating their personalized demonstrations.

\subsection{\emph{Return}: Knowing What We Don't Know} \label{sec:classify}

If the human repeats a task that the robot has seen many times before (e.g., opening the fridge), we can rely on our model to assist the human. But what happens if the human tries to perform a new or rarely seen task? Here we \textit{do not trust} the robot's assistive actions since this task is out of the robot's training distribution. In general, deciding where to arbitrate control requires a trade-off: we want the robot to take as many autonomous actions as possible (reducing the human's burden), but we don't want the assistive robot to over-commit to erroneous behavior, and prevent the human from doing what they really intended.

To solve this problem we take inspiration from recent work on safe imitation learning \cite{zhang2017query,menda2017dropoutdagger,kelly2019hg}. Our goal is to determine when the robot should trust the collective output of Equations~(\ref{eq:M1}) and (\ref{eq:M2}). Intuitively, if the human's behavior $\tau^i$ is unlike any seen behavior $\tau \in \mathcal{D}$ we should return control to the human. We therefore train a discriminator $\mathcal{C}$ that distinguishes \textit{seen} behavior from \textit{unseen} behavior. Unseen behavior is cheap to produce: we can generate this behavior by applying noisy deformations to the observed interactions $\tau \in \mathcal{D}$. At run time, our discriminator outputs a scalar confidence over the human's current behavior, which we then utilize to arbitrate control between human and robot:
\begin{equation} \label{eq:M4}
    \beta \propto \mathcal{C}(\tau^i)
\end{equation}
Recall that $\beta$ from \eq{P2} blends $a_{\mathcal{R}}$ and $a_{\mathcal{H}}$. If $\tau^i$ deviates from previously seen input patterns, $\beta \rightarrow 0$, and the human intervenes to complete this new task. By contrast, if the discriminator recognizes $\tau^i$ as similar to previous experience, $\beta \rightarrow 1$ and the robot provides assistance.

\p{Continual Learning} During an interaction the robot applies Equations~(\ref{eq:M1}), (\ref{eq:M2}), and (\ref{eq:M4}) to assist the human. But what about \textit{between} interactions? Let's say we train our encoder, decoder, and discriminator after the human has collaborated with the robot for a single week. Over the next week the human uses their robot, they will inevitably perform new tasks. An intelligent assistive robot should also learn these tasks and \textit{continuously adapt} to the human. At the end of interaction $i$, we therefore add $\tau^i$ to dataset $\mathcal{D}$. We then \textit{retrain} our models between interactions, updating $\mathcal{E}$, $\pi_{\mathcal{R}}$, and $\mathcal{C}$. Intermittent retraining enables the robot to continually learn and refine tasks over repeated interaction.
\section{Simulations}

We have proposed an algorithm that learns to assist users over repeated interaction. Our algorithm breaks down into three parts: recognizing the task, replicating prior demonstrations, and returning control when unsure. But how does our approach compare to other imitation learning baselines that also learn from repeated interaction? For instance, what if we remove our encoder, and simply train a behavior cloning agent conditioned on the current interaction $\tau^i$? In this section we perform an \textit{ablation study} in which we determine how recognizing, replicating, and returning all contribute to overall robot success. We conduct these experiments on a real robot arm with simulated human operators.

\p{Experimental Setup} We implement our approach (\textbf{Ours}) on a $7$-DoF FrankaEmika robot arm. A simulated user controls this robot to both reach \textit{discrete goals} (e.g., grasping a can) and perform \textit{continuous skills} (e.g., opening a drawer). This simulated user is not perfect: the human selects commanded actions $a_{\mathcal{H}}$ with varying levels of Gaussian white noise.

\begin{figure}[t!]
	\begin{center}
		\includegraphics[width=1.0\columnwidth]{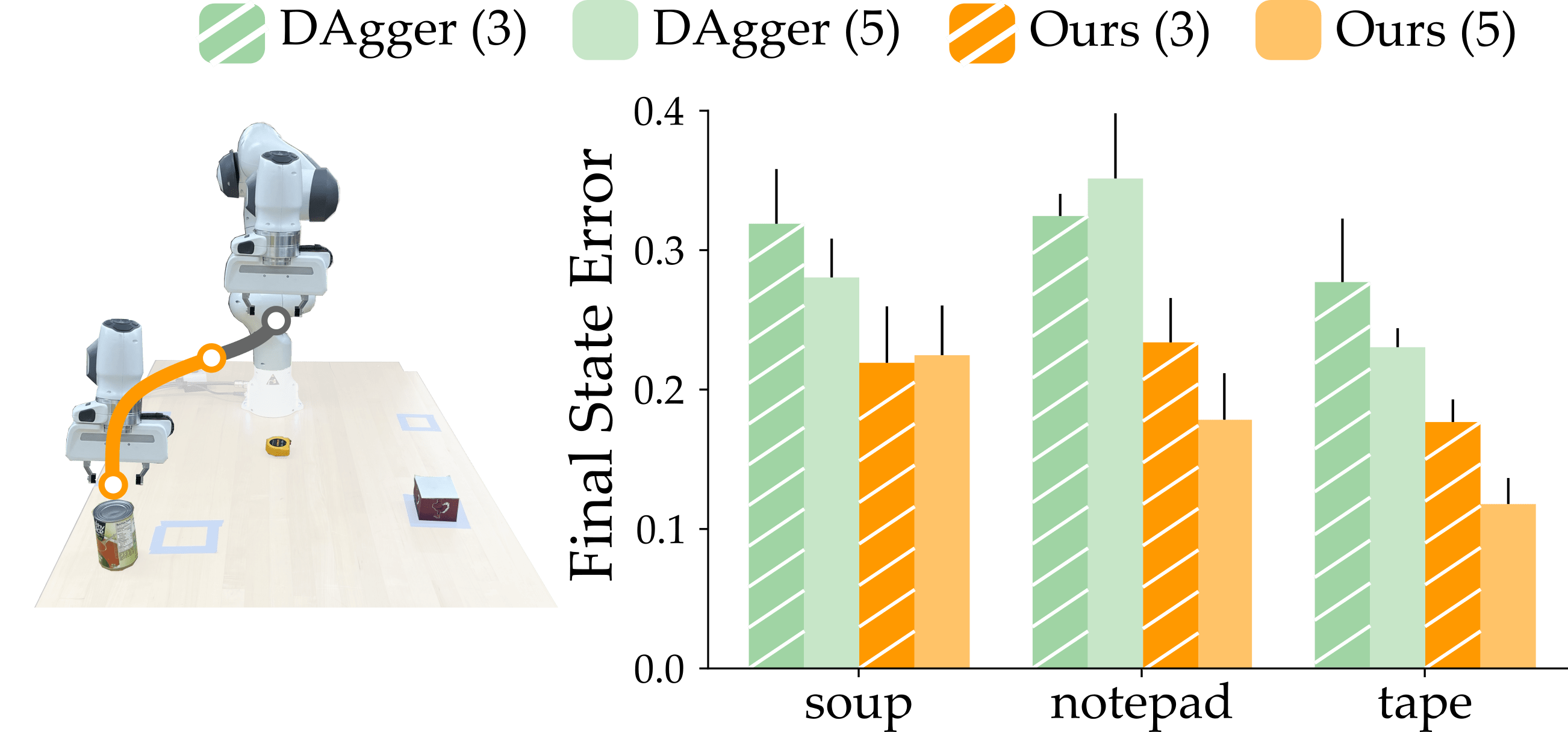}

		\caption{Comparison to \textbf{DAgger} \cite{ross2011reduction}, an imitation learning baseline that does not use latent embeddings. A simulated user controls the robot for the first $0.5$ s of the interaction: the robot must recognize the human's task and complete the rest of the reaching motion autonomously. We measure the final state error for each goal after training with $3$ or $5$ repeated interactions.}
		
		\label{fig:sim1}
	\end{center}
	\vspace{-1em}

\end{figure}

\subsection{Do We Need Recognition?}

In our first experiment with simulated humans we explore whether we need a separate module for task recognition. Recall that in Section~\ref{sec:encoder} we introduced an encoder which embeds the current interaction $\tau^i$ to latent task $z \in \mathcal{Z}$. Within Section~\ref{sec:decoder}, we then decoded $z$ into an assistive robot action using $\pi_{\mathcal{R}}(s, z)$. But it's natural to ask if we need this encoder in the first place: in other words, can we obtain similar performance \textit{without} embedding to latent space $\mathcal{Z}$? Here we consider an alternative to \eq{M2} where the robot \textit{directly} learns $\pi_{\mathcal{R}}(s, \tau^i)$ using behavior cloning. Specifically, we baseline against Dataset Aggregation (\textbf{DAgger}) \cite{ross2011reduction}.

The environment here consists of three potential goals: a can of soup, a notepad, or a tape measure. The human teleoperates the robot along $3$ or $5$ demonstrations to reach each goal. We train \textbf{Our} approach and the \textbf{DAgger} baseline from these repeated interactions. At test time, the human guides the robot for the first $0.5$~s of the task: based on this input, the robot must recognize which task the human is trying to perform and automate the rest of the reaching motion. We plot the resulting error between the human's goal and the robot's final state in \fig{sim1}. As expected, the robot is better able to automate these reaching tasks after more interactions (i.e., compare $3$ to $5$). But the robot also better identifies the human's task when using our encoder: as compared to \textbf{DAgger}, \textbf{Ours} more accurately reaches the human's goal given the same amount of training data and human input. We conclude that incorporating an encoder for task recognition \textit{does improve} the robot's assistance.

\begin{figure}[t!]
	\begin{center}
		\includegraphics[width=1.0\columnwidth]{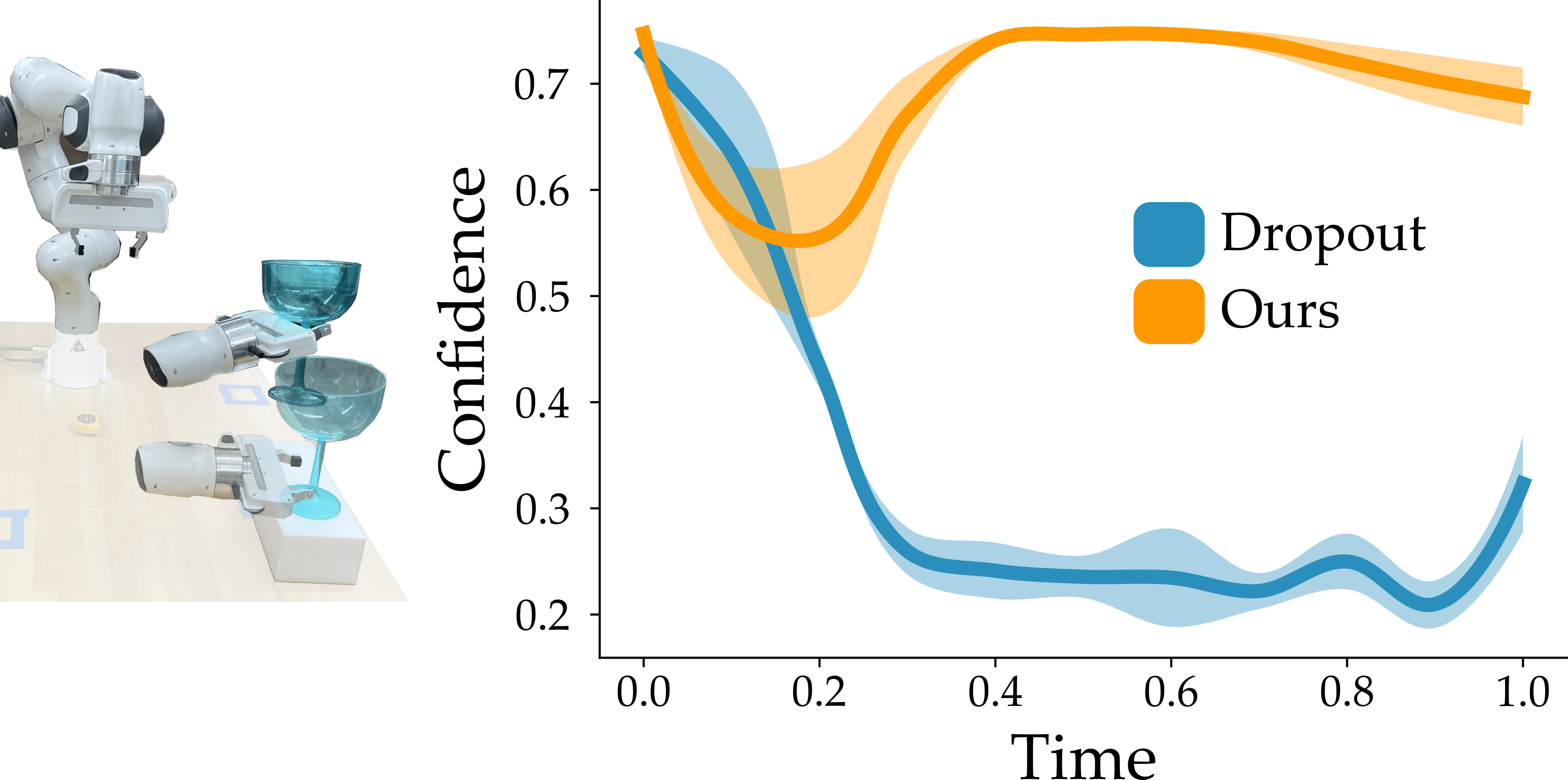}

		\caption{Comparison to \textbf{Dropout}DAgger \cite{menda2017dropoutdagger}, a safe imitation learning baseline where the robot's learned policy $\pi_{\mathcal{R}}$ evaluates its own confidence. Simulated users attempt to lift a glass. Although the robot has seen this continuous skill $5$ times before, with \textbf{Dropout} the robot is overly sensitive to minor deviations from previous interactions and rarely provides assistance.}
		
		\label{fig:sim2}
	\end{center}
	\vspace{-1em}

\end{figure}

\subsection{Do We Need Help Returning?}

In our second experiment we consider the opposite end of our pipeline: determining when the robot should provide assistance. Our intuition from Section~\ref{sec:classify} is that --- when we see an interaction similar to prior interactions --- we should trust the output of our learned model. But an alternative approach is to rely on the confidence of our learned model \textit{itself}. Here we turn to prior work on safe imitation learning where the robot samples its encoder-decoder multiple times at the current state, and assesses the similarity of the resulting actions $a_{\mathcal{R}}$. If all of these actions are almost identical, the learned model is \textit{confident} it knows what to do; conversely, if the model outputs have high variance, the robot is \textit{unsure}. Specifically, we compare DropoutDAgger \cite{menda2017dropoutdagger} (\textbf{Dropout}) to \textbf{Ours}. Recall that in \textbf{Ours} the robot trains a \textit{separate} discriminator to detect whether or not $\tau^i$ is similar to previous interactions as opposed to relying on the autoencoder model to measure its own confidence.

We compare \textbf{Dropout} and \textbf{Ours} in a continuous manipulation task where the simulated human is trying to reach and lift a glass. During test time, the human and robot share control throughout the entire interaction using \eq{P2}. The robot has seen the human perform this task five times before, and therefore should be confident in providing assistance. We visualize the robot's actual confidence $\beta$ in \fig{sim2}. Interestingly, we find that \textbf{Dropout} is overly sensitive to minor deviations from previous interactions, and incorrectly returns control to the human even when the robot can still provide useful help. \textbf{Ours} remains confident throughout this known task, suggesting that our separate discriminator better arbitrates control than the learned policy itself.

\begin{figure}[t!]
	\begin{center}
		\includegraphics[width=1.0\columnwidth]{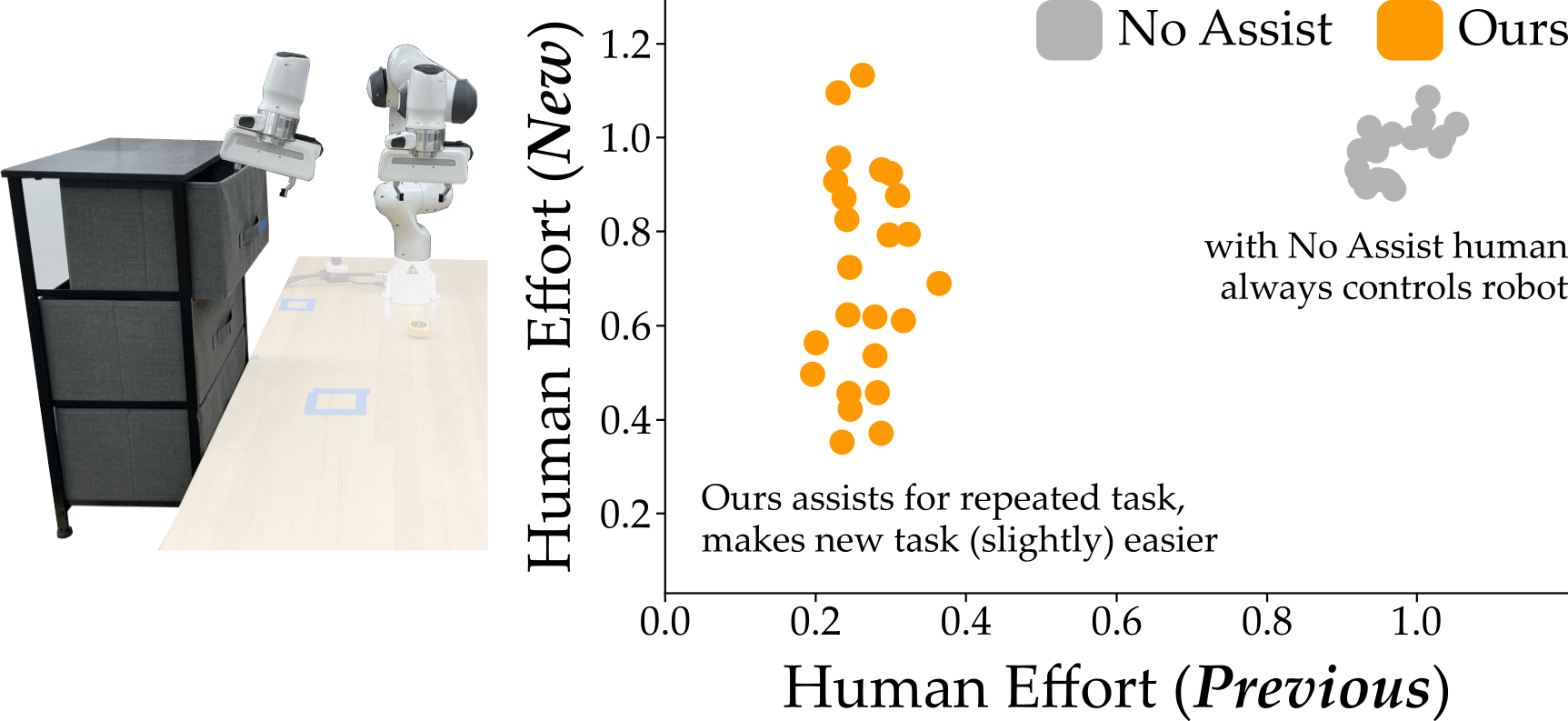}

		\caption{Simulated users alternate between a previously seen task (opening the drawer) and a new task (reaching a cup). With \textbf{No Assist}, both new and previous tasks take about the same amount of human effort. Our approach learns to partially automate the previously seen task without resisting or overriding humans when they try to complete the new task.}
		
		\label{fig:sim31}
	\end{center}
	\vspace{-0.5em}

\end{figure}

\begin{figure}[t!]
	\begin{center}
		\includegraphics[width=1.0\columnwidth]{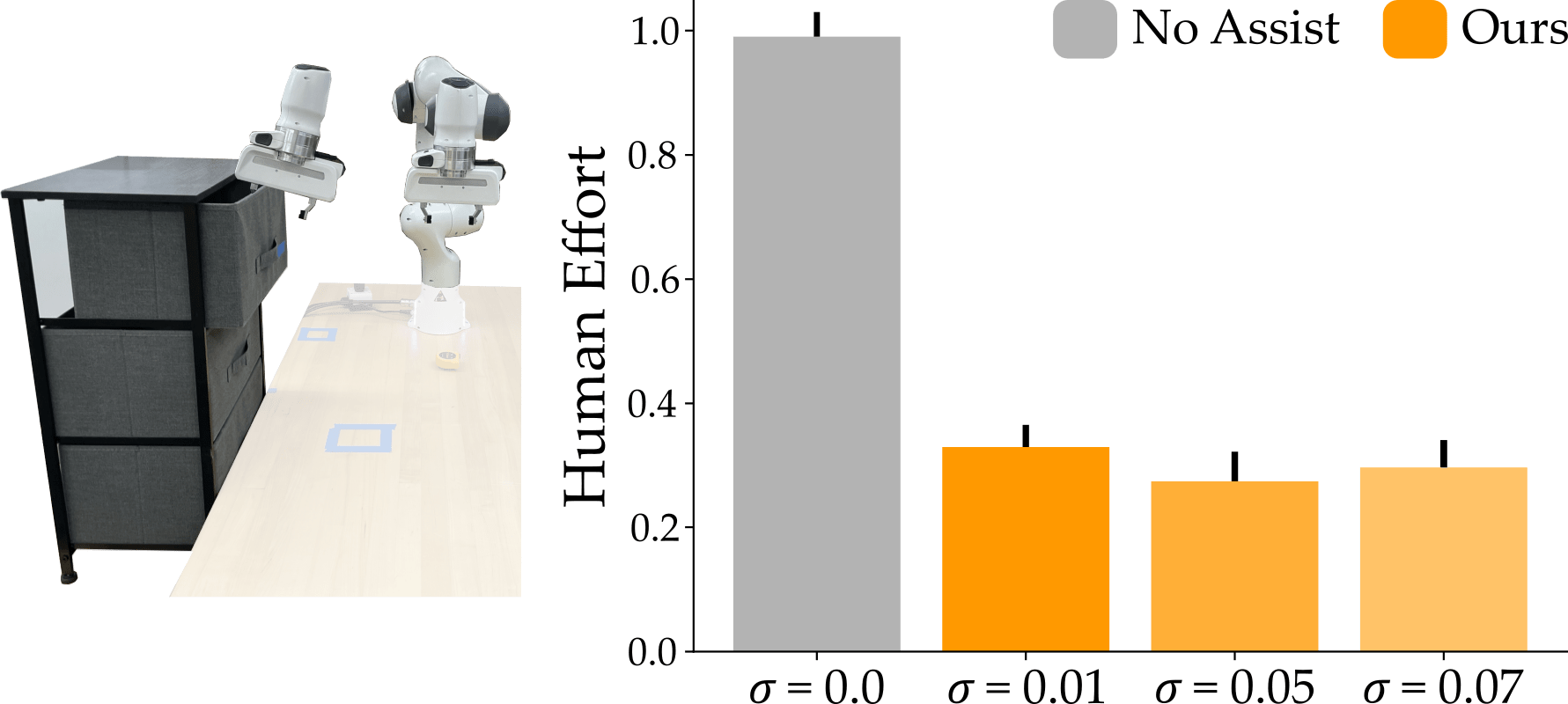}

		\caption{Simulated humans with increasingly noisy behavior. Here the human is always attempting to open the drawer (the repeated task from \fig{sim31}). We find that our robot correctly recognizes and assists for this task despite noisy and imperfect human teleoperation inputs $a_{\mathcal{H}} \sim \mathcal{N}(a_{\mathcal{H}}^*, \sigma)$.}
		
		\label{fig:sim32}
	\end{center}
	\vspace{-1.5em}

\end{figure}

\subsection{New Tasks and Noisy Humans}

So far our approach recognizes and assists the human in repeated tasks. But what about new tasks the robot has never seen before? Of course, the robot cannot assist on these tasks without any prior experience: but we want to make sure that the robot does not \textit{resist} the human, or \textit{force} them along a previously seen trajectory. This problem becomes particularly challenging when the human is noisy, and the robot must determine whether the imperfect human is trying to repeat a prior task or complete a new task.

We consider an environment that consists of a previously seen skill (opening a drawer) and an unseen goal (reaching a can). \textbf{Ours} is trained with five repetitions of the drawer skill: we compare our approach to a \textbf{No Assist} baseline, where the human teleoperates the robot without any shared autonomy. At test time, the simulated human alternates between the new and previous tasks with varying levels of Gaussian white noise ($\sigma$). Our results are shown in Figures~\ref{fig:sim31} and \ref{fig:sim32}. Here \textit{Human Effort} is the amount of time the human teleoperates the robot normalized by the average time required to complete the task. As expected, \textbf{Ours} makes it easier for the human to repeatedly open the drawer --- but on the new task, \textbf{Ours} also correctly returns control to the human. Performing the new task takes no more effort than the \textbf{No Assist} baseline; indeed, it often requires less human effort, since the robot leverages what it knows from the drawer skill to automate the start of the reaching motion. This result is robust to human noise. For the previous task, \textbf{Ours} consistently reduces human effort across multiple noise levels $\sigma$.

\section{User Study}

\begin{figure}[t]
	\begin{center}
		\includegraphics[width=1.0\columnwidth]{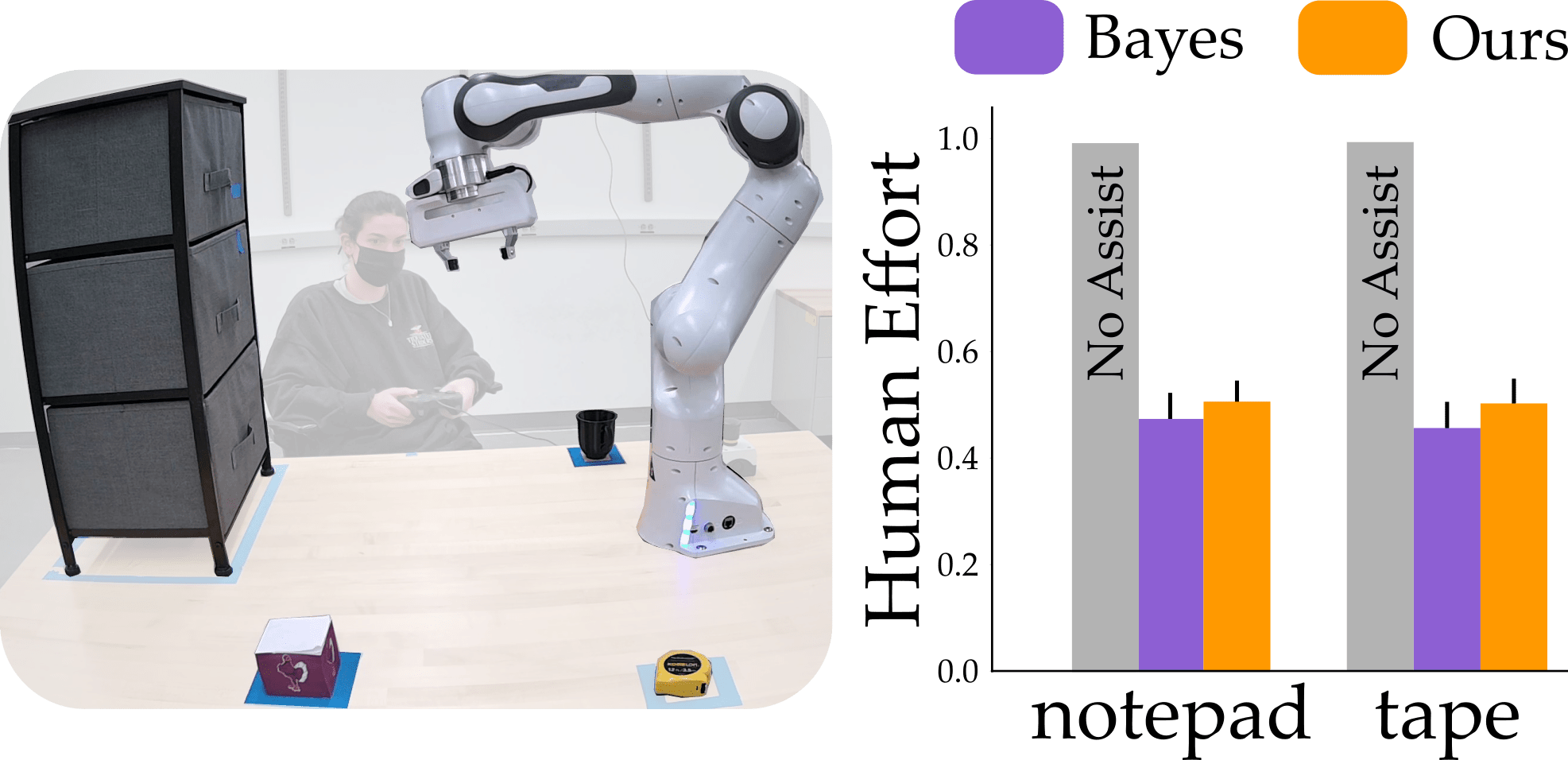}

		\caption{Experimental setup and objective results from the first part of the user study. Participants teleoperated the robot arm to reach for two goals that were known \textit{a priori}. We find that our approach (\textbf{Ours}) learns to offer assistance on par with a shared autonomy baseline (\textbf{Bayes}) \cite{dragan2013policy}. Note that we cannot leverage \textbf{Bayes} when the human wants to perform new, unexpected tasks (e.g., reaching the cup or opening the drawer), as shown in \fig{appendix}. We include the human effort under \textbf{No Assist} as a baseline.} 
		\label{fig:user1}
	\end{center}
	\vspace{-1em}

\end{figure}

Our target application is assistive robotic arms: we want to enable these arms to share autonomy with humans on everyday tasks. Motivated by this application, we conducted an in-person user study in which participants teleoperated a 7-DoF robot arm around a simplified household environment (see Figures~\ref{fig:front} and \ref{fig:user1}). We divided the study into \textit{three parts} to explore known and new tasks as well as discrete goals and continuous skills. Participants started by reaching for known goals, then taught the robot new skills, and finally returned to the original tasks. Unlike in our simulations --- where we compared our approach to imitation learning alternatives --- here we focus on current methods for shared autonomy.

\p{Independent Variables and Experimental Setup} Our user study is divided into the three sections described below. Each participant completed every section.

In the first part of the user study participants teleoperated the robot to reach for two discrete goals placed on the table. These potential goals were known \textit{a priori}, and the robot had prior experience reaching for them. Here we compare \textbf{Our} proposed approach to an existing shared autonomy baseline (\textbf{Bayes}) \cite{dragan2013policy}. \textbf{Bayes} exploits the prior knowledge about the goal set to infer which goal the human is trying to reach and assist them during the task. Similarly, \textbf{Ours} learns from the previous, offline demonstrations to recognize the user's goal and assist the human during the current task.

The shared autonomy baseline is the gold standard when the human wants to complete a task that the robot already knows --- but what happens during new tasks? In the second part of the study, participants iteratively performed two new tasks a total of $9$ times each. One task was a discrete goal (reaching a cup), while the other was a continuous skill (opening a drawer). Here we compare \textbf{Ours} to the \textbf{No Assist} baseline, where the human completes these tasks without any robot assistance. We retrained our approach every three trials for both tasks: intuitively, \textbf{Ours} should increasingly assist the user as it gets more familiar with these new tasks.

One concern with our approach is that it will specialize in just one or two recent tasks without remembering older tasks. In the last part of the user study, participants take the final learned model from both new tasks and use it to revisit the original reaching tasks. Here we compare three conditions: \textbf{No Assist}, where the human acts alone, \textbf{Ours (task)}, the robot's learned assistance with just the user's data from that specific task, and \textbf{Ours (all)}, our approach trained on the user's full dataset of interactions.

\p{Dependent Measures -- Objective} Across all three parts of the user study we measured \textit{Human Effort}. Human effort captures the amount of time the human teleoperated the robot during the task. We normalize this by the average time taken to complete the task. Higher values of human effort indicate that the human spent more time guiding the robot's motion.

\begin{figure}[t]
	\begin{center}
		\includegraphics[width=1.0\columnwidth]{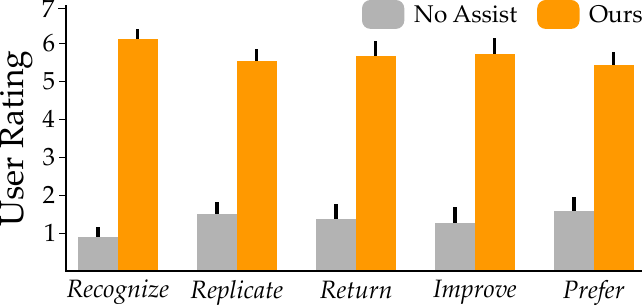}

		\caption{Subjective results from our in-person user study. Higher ratings indicate user agreement. Overall, participants thought our method provided useful assistance, and they preferred this assistance to trying to complete the tasks by themselves.}
		\label{fig:likert}
	\end{center}
	\vspace{-2em}

\end{figure}

\p{Dependent Measures -- Subjective} We administered a 7-point Likert scale survey after users completed the study (see Figure \ref{fig:likert}). Questions were organized along five scales: how confident users were that the robot \textit{Recognized} their objective, how helpful the robot's behavior was (\textit{Replicate}), how trustworthy users thought the robot was (\textit{Return}), whether the robot improved after successive demonstrations (\textit{Improve}), and if they would collaborate with the robot again (\textit{Prefer}).

\p{Participants and Procedure}
A total of $10$ members of the Virginia Tech community participated in our study ($3$ female, $1$ non-binary, average age $22 \pm 7$ years). All participants provided informed written consent prior to the experiment. 

\p{Hypotheses}
We tested three main hypotheses:
\begin{itemize}
    \item[] \textbf{H1.} \textit{In cases where the robot has prior knowledge about the human's potential goals, our learning approach will perform similarly to a shared autonomy baseline}
    \item[] \textbf{H2.} \textit{In cases where the human repeatedly performs new tasks, our approach will learn to provide meaningful assistance from scratch}
    \item[] \textbf{H3.} \textit{Our robot will remember how to assist users on previously seen tasks even after learning new ones}
\end{itemize}

\begin{figure*}[t!]
	\begin{center}
		\includegraphics[width=2\columnwidth]{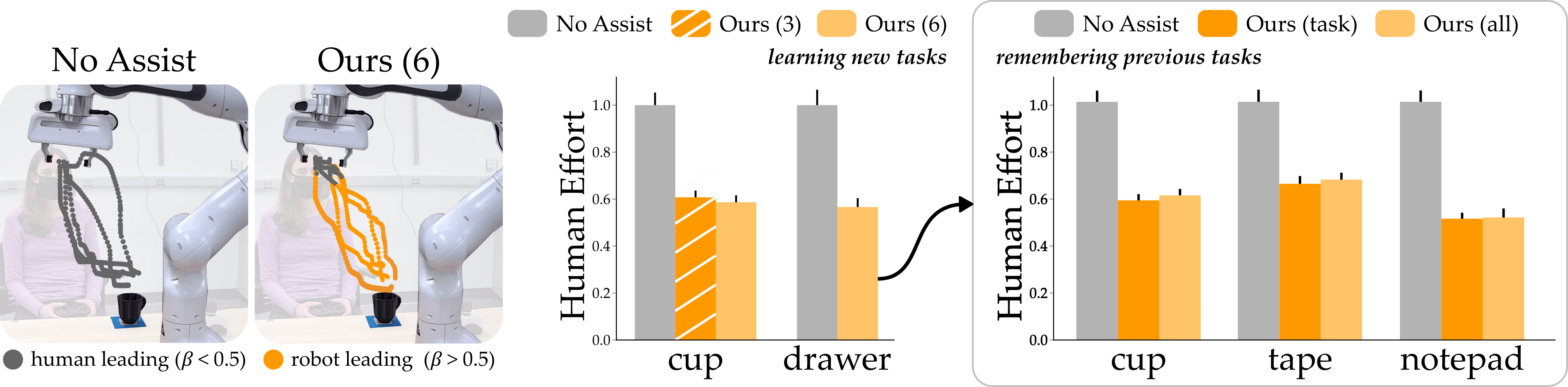}

		\caption{Objective results from the second and third parts of our user study. (Left) The human teleoperates the robot to reach for a goal it did not know about beforehand. The first few times they interact, the user must lead the robot throughout the entire task (\textbf{No Assist}). After training \textbf{Ours} on six repeated interactions, the robot recognizes the human's intent and takes the lead. (Center) Across $3-6$ repeated interactions the robot learns to provide assistance for a new goal (cup) and skill (drawer). This assistance reduces the human's effort as compared to completing the task alone. (Right) We take our resulting model trained on all user demonstrations and revisit the original tasks. \textbf{Ours (all)} offers similar assistance to \textbf{Ours (task)}, a version of our approach trained only with the user's task-specific data. These results suggest the robot can learn to assist for new tasks without forgetting older ones.}
		\label{fig:user2}
	\end{center}
	\vspace{-1em}

\end{figure*}

\p{Results} The results from each part of our user study are visualized in \fig{user1} and \fig{user2}.

In the first part of the user study participants completed a reaching task with both \textbf{Bayes} and \textbf{Ours}. We found that the differences in human effort were not statistically significant: users could reach for known, discrete goals just as easily with \textbf{Ours} as they could with the shared autonomy baseline. Note that in this task the robot had to recognize which goal the human was reaching for (i.e., either the notepad or tape) and then assist the user while reaching for that target. We emphasize that we cannot apply \textbf{Bayes} when the participant is trying to perform a \textit{new} task (such as opening the drawer), since this baseline requires prior knowledge.

In the second part of our study, the results from \fig{user2} demonstrate that \textbf{Ours} got better at providing assistance for new tasks over repeated interactions. Humans spent the most effort reaching for the cup or opening the drawer by themselves: but after training our approach on $6$ demonstrations, the robot was able to recognize the drawer skill and replicate meaningful assistance. One user commented that ``\textit{by the end I didn't provide any assistance and the robot continued to move in the correct direction}.'' We emphasize that --- throughout our entire user study --- the robot was \textit{never told} what task the participant wanted to do. Instead, the robot had to recognize the participant's current task based on that user's joystick inputs.

As a final step we trained our encoder, decoder, and discriminator from the user's demonstrations across all tasks. We found that this general \textbf{Ours (all)} performed similarly to the more specialized \textbf{Ours (task)}. For instance, when revisiting the original notepad and tape tasks from the start of the study, \textbf{Ours (all)} provided comparable assistance to \textbf{Ours (task)}.

Taken together, these results support \textbf{H1}, \textbf{H2}, and \textbf{H3}. Our approach leveraged repeated interactions to learn to share autonomy across new and old tasks with discrete goals and continuous skills. Participants generally perceived the robot's assistance as helpful. Looking at the subjective results from \fig{likert}, users thought the robot correctly recognized their intent, made the task easier to complete, and got better at providing assistance over the course of the study.

\section{Conclusion}

Today's shared autonomy algorithms often rely on prior knowledge: e.g., the robot needs to know all the human's potential tasks \textit{a priori}. Here we remove this assumption by learning to assist humans from scratch. Our approach exploits the \textit{repeated} nature of everyday human-robot interaction to recognize the human's current task, replicate similar past interactions, and return control when unsure. Our simulations and user study demonstrate that this approach extends to both discrete goals and continuous skills, and learns online from a practical number of interactions ($< 10$).

\p{Limitations} So far we have focused on how assistive robot arms can adapt to their human users. But as the robot arm gets better at sharing autonomy, the human will also \textit{co-adapt} and modify their own teleoperation strategy. For example, once the human is confident the robot recognizes their current task, the user may stop providing joystick inputs and rely on the robot entirely. A limitation of our approach is that it does not explicitly account for this co-adaptation.
\section{Appendix}

\begin{figure*}[t]
	\begin{center}
		\includegraphics[width=2.0\columnwidth]{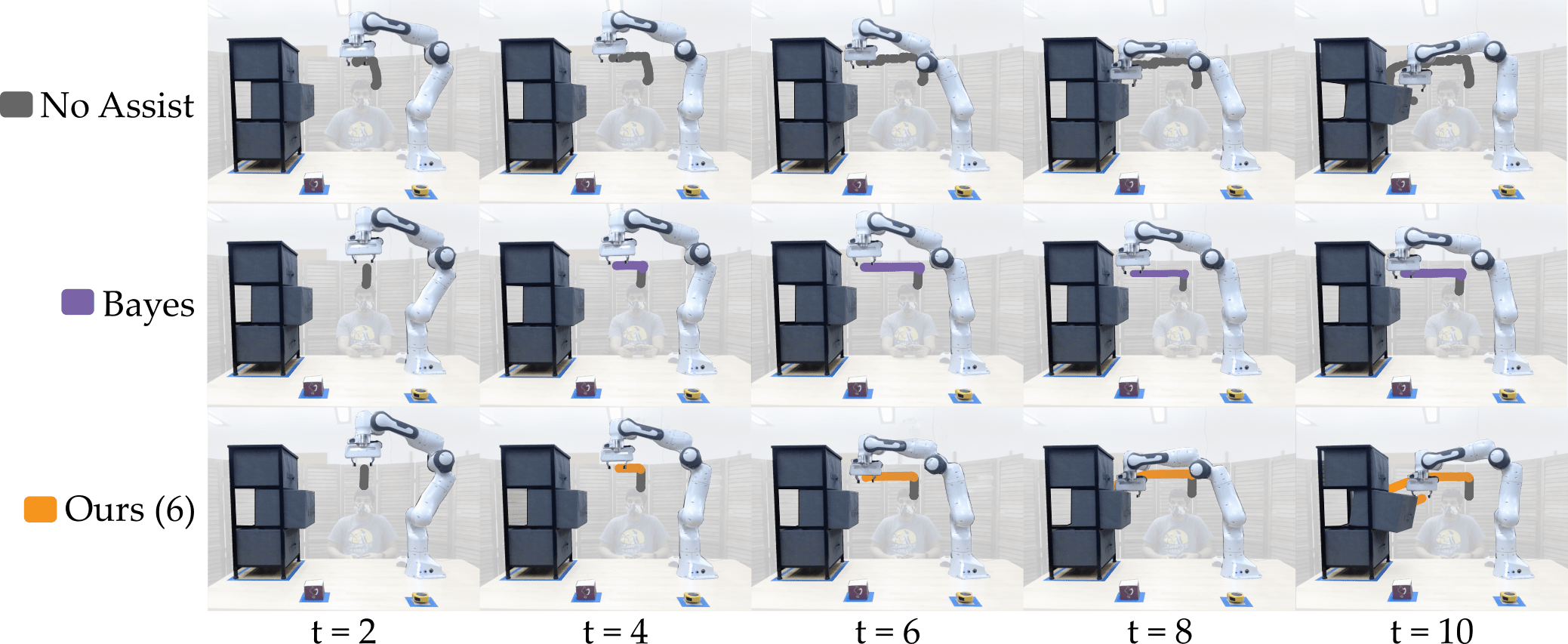}

		\caption{Representative failure case for an existing shared autonomy approach that relies on prior knowledge (\textbf{Bayes} \cite{dragan2013policy}). The user attempts to complete the drawer task with no robot assistance (\textbf{No Assist}), with \textbf{Bayes}, and with \textbf{Our} proposed method. \textbf{Bayes} has prior knowledge only about the notepad and the tape, while \textbf{Ours} has been trained on six repeated interactions for the drawer task. The user is able to successfully open the drawer by themselves (top) and with our method (bottom). With \textbf{Ours}, we see that the user is initially leading the robot towards the drawer, but once the robot recognizes this task, it takes charge and offers appropriate assistance. By contrast, \textbf{Bayes} (middle) mistakes the initial trajectory as towards the notepad, and tries leading the robot towards this known goal. Since both the drawer and the notepad are in front of the robot, the robot is initially able to move in the correct direction. However, after the user's inputs diverge from the notepad and go towards the drawer, the robot gets stuck due to conflicting commands.}
		\label{fig:appendix}
	\end{center}
	\vspace{-1.1em}
\end{figure*}

To highlight one shortcoming of state-of-the-art shared autonomy approaches and explain why \textbf{Bayes} is not a baseline in the second and third parts of our user study, we illustrate a failure case in \fig{appendix}. Here the user wants to open the drawer, but the robot only has prior knowledge about the notepad and the tape. Recall that under \textbf{Bayes} the robot infers which discrete goal the human is trying to reach and then assists towards that goal \cite{dragan2013policy, jain2019probabilistic}. But in this scenario the robot \textit{does not know beforehand} that the human may want to open the drawer. As a result, \textbf{Bayes} misinterprets the user's inputs and gradually becomes convinced that the human's target is actually the notepad \textit{next} to the drawer. This ends in a deadlock: the human teleoperates the robot towards the drawer, while the robot resists and refuses to return control. Note that the trajectories for \textbf{Ours} and \textbf{No Assist} are similar --- the main difference is that in \textbf{Ours} the robot takes the lead and automates the continuous skill.


\balance
\bibliographystyle{IEEEtran}
\bibliography{IEEEabrv,bibtex}

\end{document}